\documentclass[conference,a4paper]{IEEEtran} 
\IEEEoverridecommandlockouts
\usepackage{cite}
\usepackage{amsmath,amssymb,amsfonts}
\usepackage{textcomp}
\usepackage{xcolor}
\usepackage{float}
\usepackage{epsfig}
\usepackage{graphicx}
\usepackage{multirow}
\usepackage[normalem]{ulem}
\usepackage{sidecap}
\usepackage{algorithm}
\usepackage{bbm}
\usepackage{algorithmicx}
\usepackage{algpseudocode}
\usepackage{booktabs}
\usepackage{tabularx} 
\usepackage{caption,subcaption}
\usepackage{multicol}
\usepackage{url}

\def\Mat#1{{\boldsymbol{#1}}}

\def\BibTeX{{\rm B\kern-.05em{\sc i\kern-.025em b}\kern-.08em
    T\kern-.1667em\lower.7ex\hbox{E}\kern-.125emX}}
    
\DeclareMathOperator{\Acc}{Acc}

\newcommand{\be}{\mathbf{e}}

\newcommand{\bp}{\mathbf{p}}

\newcommand{\bx}{\mathbf{x}}

\newcommand{\sE}{\mathcal{E}}

\newcommand{\sX}{\mathcal{X}}
\newcommand{\sY}{\mathcal{Y}}

\newcommand{\bbR}{\mathbb{R}}

\DeclareRobustCommand*{\IEEEauthorrefmark}[1]{%
  \raisebox{0pt}[0pt][0pt]{\textsuperscript{\footnotesize #1}}%
}

\begin{document}

\title{Prompt-guided Scene Generation for\\3D Zero-Shot Learning}

\author{
    \IEEEauthorblockN{
    Majid Nasiri\IEEEauthorrefmark{1}, 
    Ali Cheraghian\IEEEauthorrefmark{2,5}, 
    Townim Faisal Chowdhury\IEEEauthorrefmark{3}, 
    Sahar Ahmadi\IEEEauthorrefmark{1}, 
    Morteza Saberi\IEEEauthorrefmark{4}, 
    Shafin Rahman\IEEEauthorrefmark{3}}
    \IEEEauthorblockA{
    \IEEEauthorrefmark{1}Business school, The University of New South Wales, Australia,\\
    \IEEEauthorrefmark{2}School of Engineering, Australian National University, Australia,\\
    \IEEEauthorrefmark{3}Dept. of Electrical and Computer Engineering, North South University, Bangladesh,\\
    \IEEEauthorrefmark{4} School of Computer Science and DSI, University of Technology Sydney, Australia \\
    \IEEEauthorrefmark{5} Data61, Commonwealth Scientific and Industrial Research Organisation, Australia\\majid.nasiri@unsw.edu.au, ali.cheraghian@anu.edu.au,
    \{townim.faisal, shafin.rahman\}@northsouth.edu,
    \\sahar.ahmadi@unsw.edu.au, morteza.saberi@uts.edu.au}
}


\maketitle

\begin{abstract}
Zero-shot learning on 3D point cloud data is a related underexplored problem compared to its 2D image counterpart. 3D data brings new challenges for ZSL due to the unavailability of robust pre-trained feature extraction models. To address this problem, we propose a prompt-guided 3D scene generation and supervision method that augments 3D data to learn the network better, exploring the complex interplay of seen and unseen objects. First, we merge point clouds of two 3D models in certain ways described by a prompt. The prompt acts like the annotation describing each 3D scene. Later, we perform contrastive learning to train our proposed architecture in an end-to-end manner. We argue that 3D scenes can relate objects more efficiently than single objects because popular language models (like BERT) can achieve high performance when objects appear in a context. Our proposed prompt-guided scene generation method encapsulates data augmentation and prompt-based annotation/captioning to improve 3D ZSL performance. We have achieved state-of-the-art ZSL and generalized ZSL performance on synthetic (ModelNet40, ModelNet10) and real-scanned (ScanOjbectNN) 3D object datasets.
\end{abstract}


\section{Introduction}



Modern object recognition systems based on deep learning models require extensive labelled datasets to perform better. However, in some fields, such as healthcare, or scenarios, such as COVID-19, collecting significant amounts of annotated data is tricky. To address this issue, several methods \cite{Hinton_NIPS_2009, Changpinyo_2016_CVPR, Akata_PAMI_2016, Zhang_2017_CVPR, Xian_CVPR_2017} develop Zero-Shot Learning (ZSL) methods on 2D image data by transferring the knowledge of seen data to unseen classes with the help of semantic information obtained from pre-trained object attributes or a substantial corpus of texts. Recent achievements in ZSL on image domain~\cite{pmlr-v139-radford21a} and advancements in camera technology have motivated researchers to address the ZSL problem in the 3D domain~\cite{cheraghian2019zeroshot,cheraghian2019mitigating, cheraghian2021zero} too. This task is more challenging than 2D image data because of the complex and unordered properties of 3D point cloud data~\cite{cheraghian2021zero}. Moreover, collecting 3D data, scarce objects, is complex and costly. In this paper, we aim to propose a novel ZSL method dedicated to only 3D object scenarios.

Unlike 2D image models (VGG \cite{vgg}, ResNet \cite{He2016DeepRL}), pre-trained 3D point cloud models (Pointnet++ \cite{Article2}, PointNet \cite{Article1}) cannot provide rich quality features for classification. Earlier efforts of 3D ZSL \cite{cheraghian2019zeroshot,cheraghian2019mitigating,cheraghian2020transductive} point out this problem mentioning that 3D datasets contain a limited number of classes, whereas 2D datasets include thousands of categories. For this, the 3D ZSL problem inevitably exhibits poor visual-semantic alignment, hubness, and bias. All existing works attempt to minimize those problems by proposing triplet, hubness and biasing losses in inductive and transductive settings \cite{cheraghian2019zeroshot,cheraghian2019mitigating,cheraghian2021zero}. Similarly, replacing 2D feature extraction backbones with compatible 3D models, \cite{cheraghian2020transductive} reported 2D ZSL methods' performance on 3D data. However, none of the methods mentioned above utilizes the 3D nature of the input data. In this paper, we learn different object categories based on dynamically generated 3D scenes from pre-determined language prompts \cite{tsimpoukelli2021multimodal} treated as scene captions.


\begin{figure}
\centering
\includegraphics[width=1\linewidth]{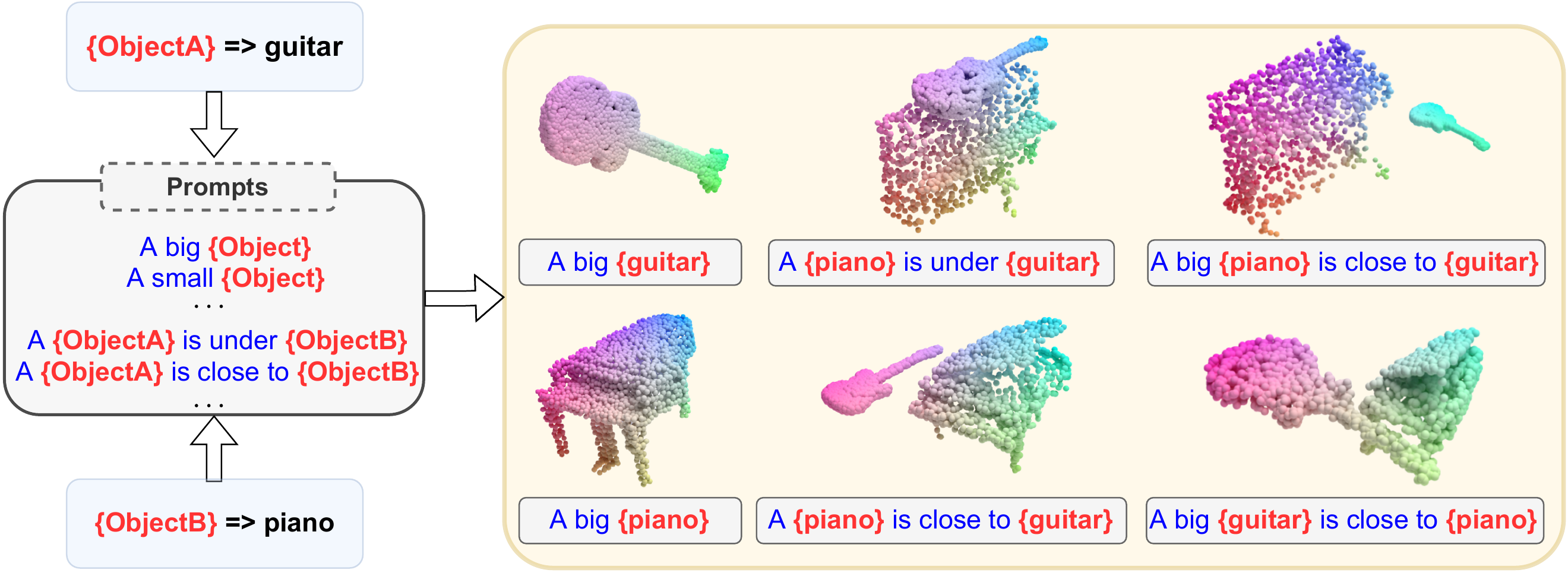}
\caption{\small Instead of the traditional use of the single point cloud model vs. single word-based semantics, we train our model based on dynamically generated 3D scenes and prompt-based captions. Here, using ObjectA (guitar), ObjectB (piano), and prompt templates (italicized sentences), we generated multiple scenes and their captions on the right.
}
\label{fig:motivation}
\end{figure}



In recent years, contrastive learning is showing promising results in combining vising and language tasks \cite{radford2021learning}. Here, language models (like BERT) can better describe a concept when a context comes into play. 2D image domain could easily take this advantage because of the availability of large-scale datasets. In contrast, since 3D point cloud datasets are not as rich as their 2D counterparts, the 3D domain cannot efficiently utilize context benefits. Moreover, obtaining 3D scene captions is a big challenge. To address this, we propose a simple but effective dynamic 3D scene generation and annotation method, particularly useful for the 3D ZSL problem. At each training iteration, following some pre-defined prompt templates (having two placeholders for two random classes), we dynamically construct a set of new 3D scenes (see Fig. \ref{fig:motivation}). A pre-trained BERT model extracts a word vector description from the prompt text, and a 3D point cloud backbone (PointNet trained from scratch) provides point cloud features. Then, a contrastive learning strategy has been applied to the entire 3D feature and word vector pairs.  Note, unlike 2D cases, we do not use a large corpus of text annotations for its difficulty in obtaining annotated captions for 3D data. Instead, we use pre-defined prompt-based captions to annotate the scene, which may not be the perfect caption. For example, `a bed is under the radio' can be a caption that may be unrealistic in natural language but included in our training. 
Nevertheless, dynamically generated scenes could provide somewhat robust context information to the network exploring detailed inter-relation of seen categories. Notably, this approach can explore more semantic relations than the traditional use of employing one 3D model and its word vector pair at a time. In this way, our method relaxes the data scarcity problem by generating many two object scenes and their prompt-based captions. It eventually improves point cloud features vs. semantic alignment to generalize the training, especially for generalized ZSL problems. We experiment with two ZSL setups using 3D point cloud datasets, ModelNet40 \cite{wu20153d}, ModelNet10\cite{wu20153d} and ScanObjectNN \cite{uy2019revisiting}, and report state-of-the-art results. The contributions of this paper are: 
\begin{itemize}
    \item dynamic 3D scene generation and prompt-based annotation methods for 3D ZSL problem,
    \item contrastive learning strategy using prompt-based 3D scene annotations,
    \item extensive experiments on both synthetic (ModelNet40, ModelNet10) and real (ScanObjectNN) datasets. Moreover, results are compared with both 2D image and 3D point cloud methods.
\end{itemize}

\section{Related work}

\noindent\textbf{Zero-shot learning:} In general, two main approaches are used to solve the ZSL problem: Embedding-based methods \cite{socher2013zero,frome2013devise,li2017zero,xian2016latent} and Generative model-based methods \cite{xian2018feature,khare2020generative,mishra2018generative,vyas2020leveraging}. In the first category, image features and semantic attributes are mapped to a shared embedding space by learning a projection function using deep networks. Frome \textit{et al.} \cite{frome2013devise} proposed a deep visual-semantic embedding model (DeViSE) to identify visual objects using both labeled image data as well as semantic information gleaned from the unannotated text. This model uses textual data to learn semantic relationships between labels and explicitly maps images into a rich semantic embedding space. Learning a linear compatibility function is not suitable for the fine-grained classification problem. Accordingly, some methods are suggested for learning nonlinear embedding. Whereas \cite{xian2016latent} learns a collection of linear models while allowing each image-class pair to choose from them. 
Generative model-based methods try to overcome the problem of bias and domain shift \cite{fu2015transductive} in the first method by generating image features for unseen classes using semantic attributes. Mishra \textit{et al.} \cite{mishra2018generative} proposed to model the statistical image generation process using a conditional variation auto-encoder (CVAE) \cite{sohn2015learning} and generate samples for unseen classes. In \cite{vyas2020leveraging}, by obtaining the semantic relationship between the seen and unseen classes by introducing Semantic Regularized Loss (SR-Loss), LsrGAN generates visual features that maintain the same semantic relationship between both classes.

\noindent\textbf{Zero-shot learning on 3D point cloud data:}
Despite the significant success of the ZSL problem in 2D images, this remains a challenging issue in 3D objects. \cite{cheraghian2019zeroshot}is a pioneering work in solving ZSL problem in 3D point cloud. In this work, the PointNet \cite{Article1} architecture is used to extract a feature space, and a bilinear compatibility function is applied to associate the  point cloud feature vector with the corresponding semantic information. Given that in \cite{cheraghian2019zeroshot}, feature space is used as embedding space, the model becomes biased in predicting a few specific labels for most of the test samples. This phenomenon is called hubness and occurs more frequently in high-dimensional data\cite{zhang2017learning}.
In \cite{cheraghian2019mitigating}, in order to solve this problem, a new loss function consisting of a regression term and a skewness term is introduced. In \cite{cheraghian2020transductive}, for the first time, a new triple loss function was introduced to solve the transductive ZSL and generalized zero-learning (GZSL) in 3d point cloud classification.This loss function uses unlabeled data in an unsupervised manner and has the ability to expand into 2D images. In \cite{michele2021generative}, unlike the methods mentioned so far, a generative ZSL method is introduced that is used for both classification and semantic segmentation in 3D objects.

\noindent\textbf{Prompt based learning:}
Prompt based learning is common unsupervised approach to train a language model for NLP tasks. GPT \cite{brown2020language, radford2019language} models are used prompt based learning to achieve zero shot and few shot performances. In classification based tasks \cite{gao2020making, schick2020exploiting}, prompt templates are reasonably easy to construct and helpful when training samples are few. Prompt-based learning is still not prevalent in computer vision tasks. Tsimpoukelli \textit{et al.} \cite{tsimpoukelli2021multimodal} used prompt augmentation technique with a fixed pretrained language model for training visual encoder model. However, in our method, we apply prompt-based learning to train a visual model to attain zero shot performance on 3D point cloud data.

\section{Method}

\begin{figure*}
\centering
\includegraphics[width=1\linewidth]{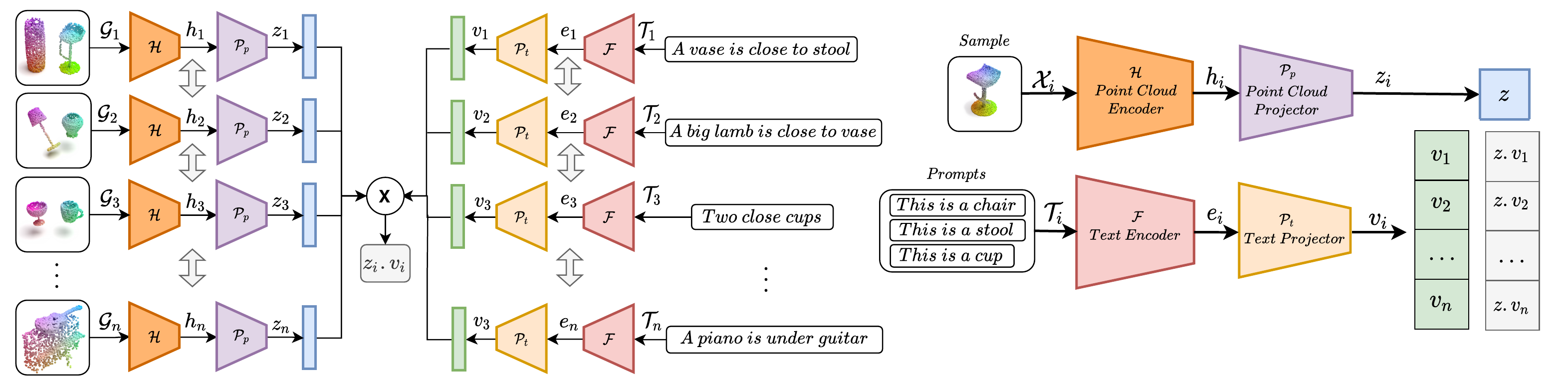}
\caption{\small Training and inference pipeline of proposed method. In the training stage, a batch of generated scenes $\mathcal{G}_{i}\in {\rm I\!R^3}$ are fed into the point cloud encoder $\mathcal{H}$ to extract the feature embedding $\textbf{h}_{i} \in {\rm I\!R^m}$, then the extract feature representation $\textbf{h}_{i}$ is forwarded into the projection module $\mathcal{P}_{p}$ in order to map into a common space, where the feature representation is $\textbf{z}_{i} \in {\rm I\!R^u}$. Similarly, for each generated scenes, there is a prompt description $\mathcal{T}_{i}$. Next, these prompt description are forwarded into a text encoder, \textit{i.e.} BERT, to obtain a feature embedding $\textbf{e}_{i}\in {\rm I\!R^d}$. After this step, the feature embedding $\textbf{e}_{i}$ is mapped into $\textbf{v}_{i}\in {\rm I\!R^u}$. After calculating $\textbf{z}_{i}$ and $\textbf{v}_{i}$ for each sample in the batch, a smility matrix is calculated for the all samples in the batch. Then, this simsilty matrix is optimized based on a loss function. In the infernce stage, a sample $\mathcal{X}^{u}$ is forwarded into the point cloud pipeline to extract $\textbf{z}_{i}$. From the language side, all prompts of seen and unseen classes are forwarded into the text pipeline, which result into the feature representation $v_{j}$. Finally, a similarity function applied between point cloud feature and prompt description of all classes to find the correct class.}
\label{fig:pipline}
\end{figure*}

\noindent\textbf{Motivation:} 
ZSL methods designed for the image domain typically benefit from strong pre-trained models, like ResNet, trained on ImageNet, which consists of millions of labeled images with thousands of categories. Thus, the extracted 2D features are nicely clustered. Nevertheless, there is no counterpart pre-trained model in the 3D domain. Labeled 3D datasets are usually small and contain only limited sets of classes. For example, the Modelnet40 dataset consists of 40 classes with only a few thousand labeled samples. As a result, the pre-trained point cloud models such as PointNet trained on Modelnet40 generate poor-quality 3D features for novel classes with clusters that are not as well-separated as their visual counterparts \cite{cheraghian2021zero}. To address this issue, we generate more 3D point cloud samples for training the pre-trained model in this paper. Specifically, we generate synthetic point cloud scenes with their associated scenes prompt from the semantic domain. More specifically, in our approach, the context and relationship between objects are also considered, which helps our model for better generalization on unseen point cloud classes. Note, in conventional ZSL approaches~\cite{socher2013zero,frome2013devise,li2017zero,khare2020generative,mishra2018generative,vyas2020leveraging}, a single object and its associated semantic class embedding are used to train a ZSL model.

\subsection{Revisiting traditional ZSL}
Suppose a 3D point cloud object instance is defined as $\sX = \{\bx_{i}\}_{i = 1}^{n}$, $ \textbf{x}_i \in {\bbR^3}$.  $\sY^{s} = \{y_{1}^{s},...,y_{S}^{s}\}$ and $\sY^{u} = \{y_{1}^{u},...,y_{S}^{u}\}$ are seen and unseen class label sets with
sizes $S$ and $U$ respectively. Here, seen and unseen labels are disjoint, i.e., $\sY^{s}\cap\sY^{u}=0$.  Additionally, $\sE^{s}=\{\phi(y_{1}^{s}), ..., \phi(y_{S}^{s})\}$ and $\sE^{u}=\{\phi(y_{1}^{u}), ..., \phi(y_{U}^{u})\}$ are the sets of semantic feature embedding for the embedding function $\phi(\cdot)$, where $\phi(y)\in\bbR^{d}$.
To this end, we define the set of $n_{s}$ seen samples as $\mathcal{D}^{s} = \{(\sX_{i}^{s}, y_{i}^{s}, \textbf{e}_{i}^{s})\}_{i=1}^{n_{s}}$, where $\sX_{i}^{s}$ is the $i$\textsuperscript{th} instance of the seen set with ground truth $y_{i}^{s} \in \sY^{s}$ and semantic vector $\be_{i}^{s} = \phi(y_{i}^{s}) \in \sE^{s}$.
Similarly, the set of $n_{u}$ unseen samples is defined as $\mathcal{D}^{u} = \{(\sX_{i}^{u}, y_{i}^{u}, \textbf{e}_{i}^{u})\}_{i=1}^{n_{u}}$, where $\sX_{i}^{u}$ is the $i$\textsuperscript{th} sample of the unseen set with ground truth $y_{i}^{u} \in \sY^{u}$ and semantic vector $\be_{i}^{u} = \phi(y_{i}^{u}) \in \sE^{u}$. The aim of ZSL in the traditional settings is to learn a prediction function  $f$ as below from the seen set $\mathcal{D}^{s}$,

\begin{equation}
f(\mathcal{X} ; \textbf{W})=\arg \max _{{y} \in \sY^{s}\cup \sY^{u}} F(\mathcal{X}, {y} ; \textbf{W}),
\end{equation}

\noindent where $F$ is a score function that ranks the correct target higher than the incorrect ones, and $\textbf{W}$ is the trainable parameters of $F$. The goal is to maximizing the performance of $F$ on test samples of seen and unseen sets. $F$ is usually takes the following form~\cite{Akata_CVPR_2013,Akata_CVPR_2015,Frome_NIPS_2013},

\begin{equation}
F(\textbf{x}, {y} ; \textbf{W})=\theta(\mathcal{X})^{\top} W \phi({y}),
\end{equation}
\noindent where $\theta(\mathcal{X})$ and $\phi({y})$ are the visual and the semantic embedding, respectively. $F$ is usually learnt
by minimizing the following loss function,

\begin{equation}
L=\frac{1}{n_{s}} \sum_{i=1}^{n_{s}} \mathcal{L}\left({y}_{i}, f\left(\mathcal{X}_{i} ; \textbf{W}\right)\right)+\gamma \Omega(\textbf{W}),
\end{equation}

\noindent where $\mathcal{L}$ is usually a cross-entropy loss function to learn the association linking the visual and the semantic domains, and $\Omega$ is the regularization term employed to restrain the complexity of the model.

\begin{table}[]
\begin{tabular}{ll}
\hline
\multicolumn{2}{c}{Prompt} \\ \hline
\multicolumn{1}{l|}{This is a \{Object\}.} & \{ObjectA\} is close to \{ObjectB\}. \\
\multicolumn{1}{l|}{A big \{Object\}.} & A big \{ObjectA\} is close to \{ObjectB\}. \\
\multicolumn{1}{l|}{A small \{Object\}.} & A small \{ObjectA\} is close to \{ObjectB\}. \\
\multicolumn{1}{l|}{Two \{Objects\}.} & \{ObjectA\} is on \{ObjectB\}. \\
\multicolumn{1}{l|}{Two close \{Objects\}.} & \{ObjectA\} is under \{ObjectB\}.\\
\hline
\end{tabular}
\caption{The set of prompts used for generating 3D point cloud scenes.}
\label{Table:prompt}
\end{table}

\subsection{Prompt-guided 3D scene generation for ZSL}

In this paper, to address the ZSL problem, instead of using directly the seen set $\mathcal{D}^{s}$, we generate a new set of samples, called synthetic scenes, to train our proposed framework. To be more accurate, the synthetic scenes are generated from the combinations of the objects in the seen set~$\mathcal{D}^{s}$. In order to generate semantically meaningful scenes, we use a set of predefined prompts $\mathcal{P} = \{\bp_{i}\}_{i = 1}^{M}$ with the size $M$, which are created manually in a prepossessing step. The generated scene samples consist of one or two samples from seen set. To be more specific, to generate new scenes $\mathcal{G}_i^s$, we linearly combine point cloud samples $\mathcal{X}_i^s$ of the seen set~$\mathcal{D}^{s}$,
\begin{multline}
\label{eqn:eq1}
\mathcal{G}_i^s = [\alpha_j * {T}(\mathcal{X}_j^s) + \beta_j] + [\alpha_k * {T}(\mathcal{X}_k^s) + \beta_k] 
\end{multline}

\noindent where ${T}$ is a random augmentation function, $\alpha$ is the scaling factor, and $\beta$ is the translation factor. It is essential to mention that  $\alpha$ and $\beta$ are defined based on the prompt semantic meaning. If a prompt consists of two objects, we down-sample the generated scene to make its number equal to the number of point of a single object. We define a set of ${\alpha_{small}, \alpha_{big}}$ which are chosen based on the prompt to change the size of the objects. To be more specific, the $\alpha_{small}$ is smaller than one, and $\alpha_{big}$ is greater than one.
At the end of this stage, we create a new training set, $\mathcal{S} = \{\mathcal{G}_{i}^{s}, \textbf{p}_{i}^{s}\}_{i = 1}^{n_{d}}$, which consists of synthetic scenes~$\mathcal{G}_{i}^{s}$ and their associated prompt description~$\textbf{p}_{i}^{s}$. In the training stage, the scene set $\mathcal{S}$ is only used. The prompts used in this paper is shown in Table~\ref{Table:prompt}.


\subsection{Language inspired contrasting learning}

The proposed architecture is shown in Figure~\ref{fig:pipline}. In the point cloud pipeline, the generated scene sample~$\mathcal{G}^{s}$ is forwarded into a point cloud encoder $\mathcal{H}$ to extract a feature embedding $\textbf{h}_{i} \in {\rm I\!R^m}$. Then, a projection function, $\mathcal{P}$, which consist of a few fully connected layers, is employed to map point cloud embedding features $\textbf{h}_{i}$ into a common space~$\textbf{z}_{i} \in {\rm I\!R^u}$. Similarly, in the text pipeline, a text encoder, $\mathcal{F}$, is used to project the text prompt $\mathcal{T}^{s}_{i}$ to an embedding space, $\textbf{p}_{i} \in {\rm I\!R^d}$, where $\textbf{p}_{i}=\phi({\mathcal{T}})$. After that, a projection function, $\mathcal{P}_{t}$ is applied to forward the feature embedding~$\textbf{p}_{i}$ into a common space $\textbf{v}_{i}\in {\rm I\!R^u}$. In order to train the proposed architecture, a minibatch of $N$ examples are randomly selected from synthetic scenes, where the contrastive prediction task on pairs of point cloud and prompt examples are derived from the minibatch, resulting in $2N$ data samples. The negative sample is not chosen explicitly. Instead, the $2(N-1)$ sample, which contains of point cloud data and prompt, in the batch are considred as the negative instances. The loss function for a positive pair $\textbf{z}_{i}$ and $\textbf{v}_{i}$ is defined as,

\begin{equation}
\label{eqn:eq_loss}
\ell_{i, j}=-\log \frac{\exp \left(\operatorname{sim}\left(\boldsymbol{z}_{i}, \boldsymbol{v}_{j}\right) / \tau\right)}{\sum_{k=1}^{2 N} \mathbbm{1}_{[k \neq i]} \exp \left(\operatorname{sim}\left(\boldsymbol{z}_{i}, \boldsymbol{v}_{k}\right) / \tau\right)}
\end{equation},

\noindent where $\operatorname{sim}(\boldsymbol{z}, \boldsymbol{v})=\boldsymbol{z}^{\top} \boldsymbol{v} /\|\boldsymbol{z}\|\|\boldsymbol{v}\|$ is denoted as the dot product between $l_{2}$ normalized $\textbf{z}$ and $\textbf{v}$ (\textit{i.e.}
cosine similarity, $\mathbbm{1}_{[k \neq i]} \in\{0,1\}$ is the indicator function, and $\tau$ denotes a temperature parameter. The total loss is calculated for all positive pairs, both $(i, j)$
and $(j, i)$, in a batch. The overall training stage is explained in Algorithm~\ref{alg:method}.

\begin{algorithm}[!t]
\caption{The proposed method}\label{euclid}
\begin{algorithmic}[1]
\Statex \textbf{Inputs:} $\mathcal{D}^s$
\Statex \textbf{Output:} A trained model to find $\hat{y}$ for all $\mathcal{X}^{u}$ in ZSL mode and $\mathcal{X}^{s} \cup \mathcal{X}^{u}$ in GZSL mode.

\Repeat
\For{$\forall \Mat{I}$ in $\mathcal{D}^s$}
    \For{sampled minibatch $\{\mathcal{X}_{i}\}_{i = 1}^{N}$ in $\mathcal{D}^s$}
        \State $\mathcal{G}_i \gets$ generate new scene using Eq~\ref{eqn:eq1}
        \State $\mathcal{T}_j \gets$ generate new prompt
        \State $\textbf{h}_i$ $\gets$ Forward sample $\mathcal{G}^s_i$ to Encoder $\mathcal{H}$
        \State $\textbf{p}_j$ $\gets$ Forward sample $\mathcal{T}_j$ to Text Encoder $\mathcal{F}$
        \State $\textbf{z}_i$ $\gets$ Forward $\textbf{h}_i$ to $\mathcal{P}_p$
        \State $\textbf{v}_j$ $\gets$ Forward $\textbf{p}_j$ to $\mathcal{P}_t$
    \EndFor
    \For{$\forall i \in \{1, ..., 2N\}$ and $j \in \{1, ..., 2N\}$}
        \State calculate loss using Eq~\ref{eqn:eq_loss} for all pairs
    \EndFor
    \State Backpropagate and update $\mathcal{H}, \mathcal{P}_p$ and  $\mathcal{P}_t$ \EndFor
\Until convergence
\Statex \textbf{ZSL evaluation stage}
\State $\mathcal{T}_{zsl} \gets$ generate "This is a \{\textcolor{red}{Object}\}." prompt for unseen classes.
\For{$\forall \Mat{I}$ in $(\mathcal{X}^u, \mathcal{Y}^u)$ from  test  part}
    \State $\Mat{z}_i$ $\gets$ $\mathcal{P}_p(\mathcal{H}(\mathcal{X}_i))$
    \State $\Mat{v}_{zsl}$ $\gets$ $\mathcal{P}_t(\mathcal{F}(\mathcal{T}_{zsl}))$
    \State \textbf{Return} $\hat{\Mat{y}}$ based on most similar $\Mat{v}_{zsl}$ to $\Mat{z}_i$ 
\EndFor
\Statex \textbf{GZSL evaluation stage}
\State $\mathcal{T}_{gzsl} \gets$ generate "This is a \{\textcolor{red}{Object}\}." prompt for seen and unseen classes.
\For{$\forall \Mat{I}$ in $(\mathcal{X}^s, \mathcal{Y}^s, \mathcal{X}^u, \mathcal{Y}^u)$  from test part}
    \State $\Mat{z}_i$ $\gets$ $\mathcal{P}_p(\mathcal{H}(\mathcal{X}_i))$
    \State $\Mat{v}_{gzsl}$ $\gets$ $\mathcal{P}_t(\mathcal{F}(\mathcal{T}_{gzsl}))$
    \State \textbf{Return} $\hat{\Mat{y}}$ based on most similar $\Mat{v}_{gzsl}$ to $\Mat{z}_i$ 
\EndFor
\end{algorithmic}
\label{alg:method}
\end{algorithm}

\begin{table}[!t]\centering\small
\newcolumntype{C}{>{\centering\arraybackslash}X}
\renewcommand*{\arraystretch}{1.1}
\setlength{\tabcolsep}{2pt}
\begin{tabularx}{\columnwidth}{l l C C c}\hline
&\multirow{2}{*}{Dataset}    & Total   & Seen/   & Train/ \\
                            & &classes  & Unseen  & Valid/Test \\ \hline
\multirow{4}{*}{}&ModelNet40 \cite{wu20153d} & 40 & 30/--         & 5852/1560/--\\
&ModelNet10 \cite{wu20153d} & 10 & --/10         & --/--/908\\
&ScanObjectNN \cite{uy2019revisiting}    & 15 & --/11         & --/--/495\\
\hline
\end{tabularx}

\caption{Statistics of the datasets. The total number of classes (3D models or images) in the datasets is listed, beside the exact splits employed in this paper separating the classes into seen or unseen and the elements into those applied for training or testing. The splits are from \cite{cheraghian2021zero}.}
\label{Table:splitting}
\end{table}

\begin{table*}[!t] \centering \small
\newcolumntype{C}{>{\centering\arraybackslash}X}
\setlength{\tabcolsep}{4pt}
\begin{tabularx}{\textwidth}{clCCCC|CCCC}\hline
\multirow{3}{*}{} & \multirow{3}{*}{Method (PointNet)} & \multicolumn{4}{c}{ModelNet10} & \multicolumn{4}{c}{ScanObjectNN} \\
\cline{3-10}
{} & {} & ZSL & \multicolumn{3}{c|}{GZSL} & ZSL & \multicolumn{3}{c}{GZSL} \\
\cline{3-10}
 &  & Acc & $\Acc_{s}$ & $\Acc_{u}$ & HM & Acc & $\Acc_{s}$ & $\Acc_{u}$ & HM  \\ 
 \hline
& DEM \cite{Zhang_2017_CVPR} & 19.2 & 76.0 & 7.2 & 13.1 & 14.8 & 78.9 & 3.2 & 6.2\\  
& LATEM \cite{latem-cvpr16} & 10.9 & - & - & - & 9.5 & - & - & - \\ 
& GFZSL \cite{GFZSL17} & 13.1 & 80.8 & 4.1 & 7.8 & 17.9 & \textbf{83.6} & 3.6 & 6.9\\
& SYNC \cite{Changpinyo_2016_CVPR} & 15.5 & - & - & - & 13.3 & - & - & -  \\ 
& GDAN \cite{gdan-cvpr19} & - & \textbf{82.1} & 0.6 & 1.2 & - & 82.8 & 0.2 & 0.4\\  
& f-CLSWGAN \cite{Xian_2018_CVPR} & 30.0 & 20.4 & 14.6 & 17.0 & 18.6 & 21.5 & \textbf{18.7} & 20.0\\
& CADA-VAE \cite{Schonfeld_2019_CVPR} & 23.0 & 79.3 & 2.6 & 5.1 & 15.1 & 80.5 & 1.0 & 2.0 \\
& GXE \cite{cvcZSL19} & 19.7 & 58.6 & 17.0 & 26.4 & 13.8 & 31.0 & 11.3 & 16.5\\
& ZSL-3D \cite{cheraghian2021zero} & 21.3 & 79.4 & 3.7 & 7.2 & 18.9 & 75.1 & 3.6 & 6.8\\
& Ours & \textbf{40.9} & 67.1 & \textbf{17.1} &  \textbf{28.0} & \textbf{24.8} & 70.6 & 14.1 & \textbf{23.5} \\
\hline
\end{tabularx}
\vspace{.2em}
\caption{ZSL and GZSL results on ModelNet10 \cite{wu20153d} and ScanObjectNN \cite{uy2019revisiting} datasets}
\label{table:GZSL_3D_pointnet}
\end{table*}

\section{Experiment}

\subsection{Experimental Setup}


We evaluate our proposed approach on three 3D datasets, ModelNet40 \cite{wu20153d}, ModelNet10, and ScanObjectNN \cite{wu20153d}. Here, ModelNet10 is a part of the 3D synthetic ModelNet40 dataset. Unlike synthetic data of ModelNet40, ScanObjectNN consists of 3D real-world point cloud data with background noise. The statistics of the datasets are shown in Table \ref{Table:splitting}. We employ two different experimental setups, including both synthetic and real-world data. Both experimental setups are previously introduced by Cheraghian \textit{et al.} \cite{cheraghian2021zero}. The first experimental setup is generated using only synthetic data. The seen classes are the 30 classes of ModelNet40 that do not appear in ModelNet10, and the unseen classes are the remaining 10 ModelNet10 classes. The second experimental setup is more challenging and realistic than the first. This configuration utilizes 26 ModelNet40 classes as seen and 11 ScanObjectNN classes as unseen classes. This is a more practical setup because we can collect many synthetic examples of seen objects during training. Nonetheless, the model may confront many real-world 3D data instances of both seen and unknown classes at test time.

\noindent\textbf{Semantic features:} For the semantic features, we employ the 768-dimensional BERT\cite{devlin2018bert} vectors for all datasets. In the non-prompt setting, we used class name or joint class names ``$objectA\ objectB$" (when generating scene including two objects) to extract feature vectors. However, in the prompt setting, we use the generated scene description to extract feature vectors.



\noindent\textbf{Evaluation metric:} We calculate the method's performance using the top-$1$ accuracydetails. In ZSL, we predict the label of an unseen class using only the unseen class's label set. In generalized ZSL (GZSL), however, we predict the class label based on both seen and unseen class labels. We additionally report the Harmonic Mean (HM) \cite{Xian_CVPR_2017} of the accuracy of the seen and unseen classes in GZSL to identify the methods that are less biased towards the seen classes without compromising the performance of both seen and unseen classes.

\begin{align}
\textrm{HM} = \frac{2 \times acc_{s} \times acc_{u} }{acc_{s} + acc_{u}}
\end{align}
\noindent
where $acc_{s}$ and $acc_{u}$ are seen and unseen class top-$1$ accuracies respectively.

\begin{table*}[!t]
\centering
\small
\begin{tabular}{c|ccccccccccc}
\midrule
\multicolumn{1}{c|}{ModelNet10} & bathtub & bed & chair & desk & dresser & monitor & night stand & sofa & table & toilet & avg \\ 
\hline
\multicolumn{1}{c|}{baseline (A) } & 0.0 & 6.0 & 66.1 & 8.1 & \textbf{73.3} & \textbf{72.0} & 39.5 & \textbf{22.0} & 0.0 & 45.0 & 33.2 \\
\multicolumn{1}{c|}{Ours (D)} & \textbf{2.4} & \textbf{60.5} & \textbf{92.9} & \textbf{28.3} & 59.1 & 35.1 & \textbf{40.8} & 7.2 & 0.0 & \textbf{83.3} & \textbf{40.9} \\
\bottomrule
\end{tabular}%
\caption{Per-class score evaluation on ModelNet10 \cite{wu20153d} dataset.}
\label{tab:perclass_modelnet}%
\end{table*}%

\begin{table*}[!t]
\centering
\small
\begin{tabular}{c|cccccccccccc}
\midrule
\multicolumn{1}{c|}{ScanObjectNN} & cabinet & chair & desk & display & door & shelf & table & bed & sink & sofa & toilet & avg\\ 
\hline
\multicolumn{1}{c|}{Baseline} & \textbf{6.7} & \textbf{49.6} & 20.0 & 0.0 & 37.6 & 42.9 & \textbf{14.8} & \textbf{28.1} & 0.0 & 14.3 & 18.0 & 21.1\\
\multicolumn{1}{c|}{Ours} & 0.0 & 0.0 & \textbf{49.4} & 0.0 & \textbf{93.3} & \textbf{57.7} & 2.5 & 6.1 & 0.0 & \textbf{25.6} & \textbf{38.1} & \textbf{24.8} \\
\bottomrule
\end{tabular}%
\caption{Per-class score evaluation on ScanObjectNN \cite{uy2019revisiting} dataset.}
\label{tab:perclass_scanobjectnn}%
\end{table*}%
\noindent\textbf{Validation strategy:} We establish a validation strategy to find the scaling factors of $\alpha_{small}$ and $\alpha_{big}$ for the scene generation stage. We randomly split the seen classes of an experimental setup into 80\% for seen validation classes and the rest of the 20\% classes for unseen validation. The grid search method is then employed to find the right hyperparameters. In the scene generation process, we find three set of $\alpha$ scaling factors to create ``small" or ``big" objects $\{(\alpha_{small}=0.2, \alpha_{big}=5), (\alpha_{small}=0.3, \alpha_{big}=3), (\alpha_{small}=0.5, \alpha_{big}=2), (\alpha_{small}=0.7, \alpha_{big}=1.5)\}$ to assign ``small" or ``big" attribute to objects in scene.

\noindent\textbf{Implementation details \footnote{Codes and models are available at: \url{https://tinyurl.com/yfh63ny6}}:} We trained our models on a single P100 GPU. Each training epoch took about one hour for our ZSL model using the hyper-parameters described in the paper. We trained models for a total of 100 epochs. We used the Adam \cite{DBLP:journals/corr/KingmaB14} optimizer  with $\beta_{1}$ = 0.9, $\beta_2$ = 0.999 and $\epsilon={10}^{-8}$. We varied the learning rate over the course of training, started with $lr={10}^{-3}$ and decreased it by $0.5$ factor every 20 epochs. Throughout this paper, we used PointNet \cite{Article1} as point cloud encoder $\mathcal{H}$ to extract 1024 dimensional $\textbf{h}_i$ features. In addition, we added two fully connected layers on top of the encoder to project features to 512, then 128-dimensional representations $\textbf{z}_i$. Also, we added a ReLU activation function after the first dense layer. For the text pipeline, we used BERT \cite{devlin2018bert} as the text encoder $\mathcal{F}$ to encode generated prompts, following two fully connected layers with 1024 and 512 output dimensions, then a ReLU and another dense layer to provide 128-dimensional representations $\textbf{v}_i$. The text encoder is kept frozen during training, similar to \cite{tsimpoukelli2021multimodal}. We use the \textit{PyTorch} framework to perform our experiments.


\begin{figure*}[!t]
\centering
\includegraphics[width=1\linewidth]{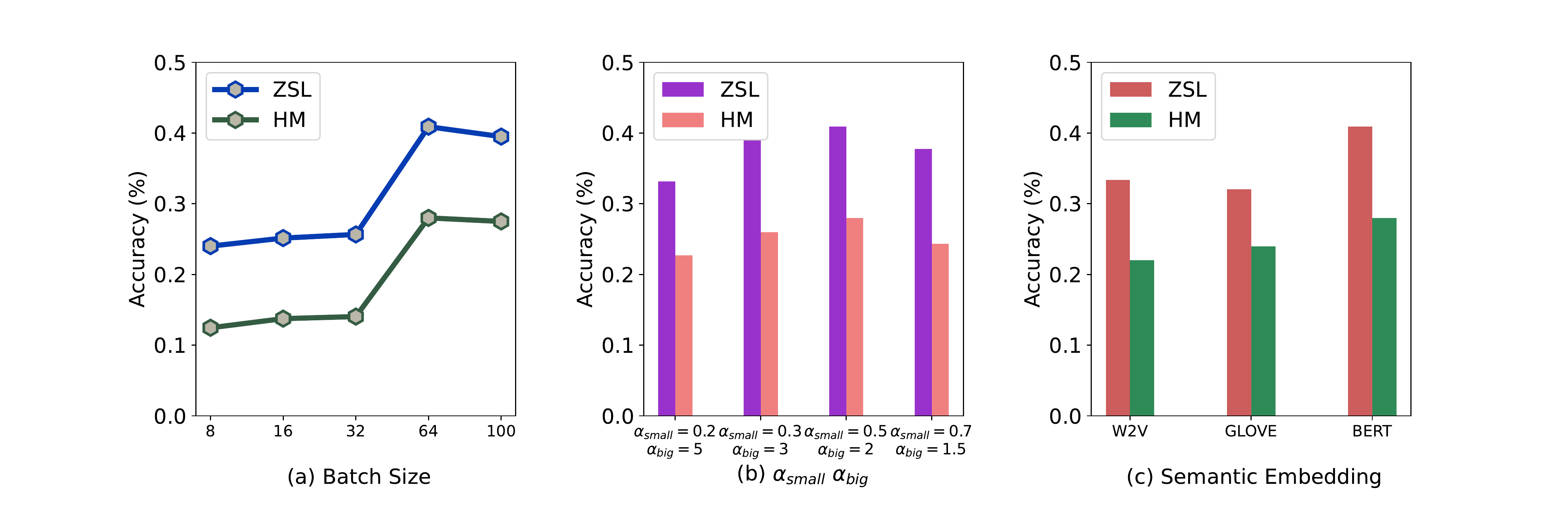}
\caption{\small The impact of different hyper-parameters in our proposed architecture. a) The effect of different batch size. b) the impact of scaling factor in the scene generation process. c) The effect of various semantic class embeddings.}
\label{fig:abliation}
\end{figure*}

\subsection{Main results}

The result of our proposed method on ModelNet10\cite{wu20153d} and ScanObjectNN~\cite{uy2019revisiting} datasets are shown in Table~\ref{table:GZSL_3D_pointnet}. We compared our proposed approach with several 2D ZSL methods (DEM \cite{Zhang_2017_CVPR}, SYNC \cite{Changpinyo_2016_CVPR}, LA\-TEM \cite{latem-cvpr16}, GDAN \cite{gdan-cvpr19}, f-CLSWGAN \cite{Xian_2018_CVPR}, CADA-VAE \cite{Schonfeld_2019_CVPR}, GXE \cite{cvcZSL19}) as well as one 3D ZSL method \cite{cheraghian2021zero}. For a fair comparison, we re-implemented these state-of-the-art 2D ZSL methods with modification of the 3D point cloud. Our proposed approach significantly beats the other methods on the ModelNet10 and ScanObjectNN datasets. 

\noindent\textbf{ModelNet10:} For the ModelNet10 dataset, as can be seen in Table~\ref{table:GZSL_3D_pointnet}, our method significantly outperform other well-known methods proposed for 2D and 3D ZSL. For ZSL, we achieve nearly more than 10\% improvement in comparison to the accuracy of the second-best method in both 2D and 3D methods. Similarly, for GZSL, we obtain the best harmonic mean 28.0\% compared to the generative 2D methods that generate samples for unseen classes and are well-known for having fewer bias issues than traditional methods. This observation shows that ZSL on 3D data is more challenging than 2D images and the current approach proposed for 2D images maybe are not best fit for this problem. In addition, we observe that the 2D methods are more biased toward the seen classes, which is why the harmonic mean is relatively low for many 2D methods. We outperform the current 3D ZSL method regarding the accuracy of unseen classes in the ZSL and GZSL setups. As a result, we design a ZSL method specific to 3D point cloud data, which can take advantage of this kind of data.

\noindent\textbf{ScanObjectNN:} In this dataset, we also outperform other state-of-the-art compared methods. Though, it is essential to notice that methods typically operate better on the 3D synthetic dataset (ModelNet10) than real data (ScanObjectNN). The reason can be the domain shift from synthetic to real data and noise in real data. As we can see in Table \ref{table:GZSL_3D_pointnet}, in ZSL, we obtain 24.8\% accuracy for predicting unseen classes, which is by a large margin better than other reported methods. We can see a similar pattern for GZSL, which shows that our proposed method can address the bias problem better than other methods.


\noindent\textbf{Per-class results:} Table \ref{tab:perclass_modelnet} and Table \ref{tab:perclass_scanobjectnn} represent the performance of each individual class from ModelNet10 and ScanObjectNN respectively. The results of both datasets are compared to a baseline method that does not include scene generation and prompt in its training pipeline. For ModelNet10, or ModelNet10, we see that our solution achieves greater than 30\% accuracy in 6 of the ten classes (bed, chair, dresser, monitor, nightstand, toilet). However, the baseline and our method do not classify instances from the table. While the baseline method mostly predicts dresser or monitor classes due to the hubness problem \cite{Shigeto_Hubness_2015,Zhang_2017_CVPR}.
On the other hand, our method can classify a few instances of the bathtub while the baseline cannot. The hubness issue in the ScanObjectNN is more severe since the real-world 3D data is noisier and more unordered than synthetic data, and there is an additional domain shift between synthetic and real data. However, our proposed method performs relatively better than the baseline method.

\subsection{Ablation study}

\noindent\textbf{Impact of batch size:} Our proposed method has a remarkable influence on the batch size. In Figure \ref{fig:abliation}(a), we report ZSL ($acc_u$) and GZSL (HM) performance on ModelNet10 using different batch sizes. We can observe that increasing batch size improves the performance. When the batch size is 64, the accuracy is at its peak for both ZSL and GZSL, and increasing the batch size to 100 has no effect on performance.

\noindent\textbf{Impact of scene generation parameters:} In this part, we evaluate the impact of scaling factor $\alpha$ in the scene generation module. As can be seen in Figure \ref{fig:abliation}(b), we achieve the best accuracy in both ZSL and GZSL when we use $\alpha_{small}=0.5$ and $\alpha_{big}=2$.

\noindent\textbf{Impact of different semantic embedding:} We have shown in Figure \ref{fig:abliation}(c) the performance of our proposed method on different semantic embedding, BERT \cite{devlin2018bert}, w2v \cite{Mikolov_NIPS_2013}, and GloVe \cite{Jeffrey_Glove_2014}. Based on this experiment, we observe that BERT achieves the best performance compared to other semantic embeddings. The reason is that the BERT model can take advantage of the context information in a prompt, while w2v and GloVe do not consider context knowledge. It is needed to mention that the w2v and GloVe feature representations of prompts are obtained by averaging among all single words in a prompt.  


\begin{table}[!t]
\centering
\small
\begin{tabular}{c|ccc|cccc}
\midrule
\multirow{2}{*}{Method} & \multirow{2}{*}{SG} & \multirow{2}{*}{P} & \multirow{2}{*}{DC} & \multirow{2}{*}{ZSL} & \multicolumn{3}{c}{GZSL} \\
 &  &  &  &  & $Acc_{s}$ & $Acc_{u}$ & HM \\ 
\hline
\multicolumn{1}{c|}{A} &  &  &  & 33.2 & \textbf{73.2} & 13.9 & 23.3 \\
\multicolumn{1}{c|}{B} & \checkmark &  & \checkmark & 34.6 & 59.3 & 16.6 & 26.0 \\
\multicolumn{1}{c|}{C} & \checkmark & \checkmark &  & 38.7 & 72.1 & 15.5 & 25.5 \\
\multicolumn{1}{c|}{D} & \checkmark & \checkmark & \checkmark & \textbf{40.9} & 67.1 & \textbf{17.1} & \textbf{28.0}\\
\bottomrule
\end{tabular}%
\caption{The impact of different component in our method: \textbf{SG}: scene generation, \textbf{P}: using prompt, in non-prompt scenario we just use categorical class labels, \textbf{DC}: generate scenes using instances of different classes vs single class in each scene}
\label{tab:addlabel}%
\end{table}%
\noindent\textbf{Impact of scene generation and prompt:} Our proposed scene generation algorithm consists of three main components: 1) Scene Generation (SG), 2) Prompt generation (P) as the class semantic information, and 3) Between class scene generation (DC). The impact of these components is reported in Table \ref{tab:addlabel}. Method A represents a vanilla model without using the three mentioned components. It is seen that the baseline method gets the lowest performance among all methods with a significant biased towards seen classes. Generating scenes using instances of different or same classes (Method B) can alleviate this bias problem of seen classes. As can be seen, our proposed method (Method D) achieves the best performance while employing all three components together.

\section{Conclusion}

Collecting 3D point cloud data of objects at a massive scale has become more accessible than ever, thanks to better 3D capture systems. Nevertheless, 3D point cloud recognition systems are not able to manage this large-scale scenario. To this end, in this paper, we propose a novel zero-shot learning framework specific for 3D point cloud data to classify previously unseen data. Our proposed approach introduces a dynamic 3D scene generation and supervision method that generates 3D scene data to learn the network better, exploring the complicated interplay of seen and unseen objects. Also, in our approach, to describe the generated scenes,  we employ a prompt-based annotation. After that, to train the proposed architecture, we perform contrastive learning. We show that 3D scenes can link objects more efficiently than single objects thanks to transformer language models (BERT). Our proposed dynamic scene generation approach encapsulates data augmentation and prompt-based annotation/captioning to enhance 3D ZSL performance. We have obtained state-of-the-art ZSL and generalized ZSL performance on synthetic (ModelNet40, ModelNet10) and real-scanned (ScanOjbectNN) 3D object datasets.

{
\bibliographystyle{IEEEtran}
\bibliography{egbib}
}

\end{document}